\documentclass[letterpaper]{article} % DO NOT CHANGE THIS
\usepackage{aaai2026}  % DO NOT CHANGE THIS
\usepackage{times}  % DO NOT CHANGE THIS
\usepackage{helvet}  % DO NOT CHANGE THIS
\usepackage{courier}  % DO NOT CHANGE THIS
\usepackage[hyphens]{url}  % DO NOT CHANGE THIS
\usepackage{graphicx} % DO NOT CHANGE THIS
\usepackage{natbib}  % DO NOT CHANGE THIS AND DO NOT ADD ANY OPTIONS TO IT
\usepackage{caption} % DO NOT CHANGE THIS AND DO NOT ADD ANY OPTIONS TO IT
\usepackage{algorithm}
\usepackage{algorithmic}

\newcommand{\model}{\textsc{GM-PRM}\xspace}
\usepackage{xspace}
\usepackage{pifont}
\usepackage{microtype}
\usepackage{amsmath}
\usepackage{tcolorbox}
\usepackage{bm}
\usepackage{amssymb}
\usepackage{booktabs}
\usepackage{array}
\usepackage{tabularx}
\usepackage{multirow}

\definecolor{green2}{RGB}{90, 130, 60}
\definecolor{gray2}{RGB}{169, 169, 169}
\definecolor{red2}{RGB}{178, 34, 34}

\newlength\savewidth
\newcommand\shline{\noalign{\global\savewidth\arrayrulewidth
                            \global\arrayrulewidth 0.8pt}%
                   \hline
                   \noalign{\global\arrayrulewidth\savewidth}}

\usepackage{newfloat}
\usepackage{listings}
\DeclareCaptionStyle{ruled}{labelfont=normalfont,labelsep=colon,strut=off} % DO NOT CHANGE THIS
\floatstyle{ruled}
\newfloat{listing}{tb}{lst}{}
\floatname{listing}{Listing}
\title{\model: A Generative Multimodal Process Reward Model for \\Multimodal Mathematical Reasoning}
\author{
    Jianghangfan Zhang\textsuperscript{\rm 1}, 
    Yibo Yan\textsuperscript{\rm 1,2}, 
    Kening Zheng\textsuperscript{\rm 1}, 
    Xin Zou\textsuperscript{\rm 1}, 
    Song Dai\textsuperscript{\rm 1}, 
    Xuming Hu\textsuperscript{\rm 1,2} \thanks{ Corresponding author.} \\
}
\affiliations{
    \textsuperscript{\rm 1}Hong Kong University of Science and Technology (Guangzhou)\\ \textsuperscript{\rm 2}Hong Kong University of Science and Technology \\
    \{zhangjhf25, yanyibo70\}@gmail.com;  \{xuminghu\}@hkust-gz.com
}

\usepackage{bibentry}
\begin{document}

\maketitle

\begin{abstract}
Multimodal Large Language Models (MLLMs) demonstrate remarkable capabilities but often \textit{struggle with complex, multi-step mathematical reasoning}, where minor errors in visual perception or logical deduction can lead to complete failure. While Process Reward Models (PRMs) offer step-by-step supervision, existing multimodal PRMs are \textit{limited to being binary verifiers that can identify but not correct errors}, offering little explanatory power. To address these deficiencies, we introduce the \textbf{Generative Multimodal Process Reward Model (\model)}, a novel paradigm that transforms the PRM from a passive judge into an active reasoning collaborator. Instead of a simple scalar score, \model provides a fine-grained, interpretable analysis of each reasoning step, evaluating its step intent, visual alignment, and logical soundness. More critically, \model is trained to generate a corrected version of the first erroneous step it identifies. This unique corrective capability enables our new test-time inference strategy, Refined Best-of-N (Refined-BoN). This framework actively enhances solution quality by using the PRM’s generated correction to guide the policy model toward a more promising reasoning trajectory, thereby improving the diversity and correctness of the solution pool. We demonstrate that \model achieves state-of-the-art results on multiple multimodal math benchmarks, significantly boosting policy model performance with remarkable data efficiency, requiring only a 20K-sample training dataset. Our code will be released upon acceptance.
\end{abstract}


\section{Introduction}
\label{sec:intro}
The advent of Multimodal Large Language Models (MLLMs) has marked a significant milestone in artificial intelligence, demonstrating remarkable capabilities in integrating and understanding visual and textual information \cite{caffagni2024revolution,yan2024urbanclip,yan2024georeasoner,huo2024mmneuron,zheng2024reefknot}. While these models excel at general-purpose tasks such as image captioning and visual question answering, they often falter when confronted with complex, multi-step reasoning challenges, particularly within specialized domains like mathematics \cite{wang2024measuring,yan2024survey,yan2025position,ahn2024large}. Solving multimodal mathematical problems requires not only accurate perception of visual elements (\textit{e.g.,} geometric figures, function graphs) but also a rigorous, step-by-step logical deduction process \cite{shi2024math,zhuang2025math,yan2025mathagent}. \textit{Minor errors in either image interpretation or logical inference can cascade, leading to entirely incorrect final answers}.

\begin{figure}[tb!]
  \centering
  \includegraphics[width=0.48\textwidth]{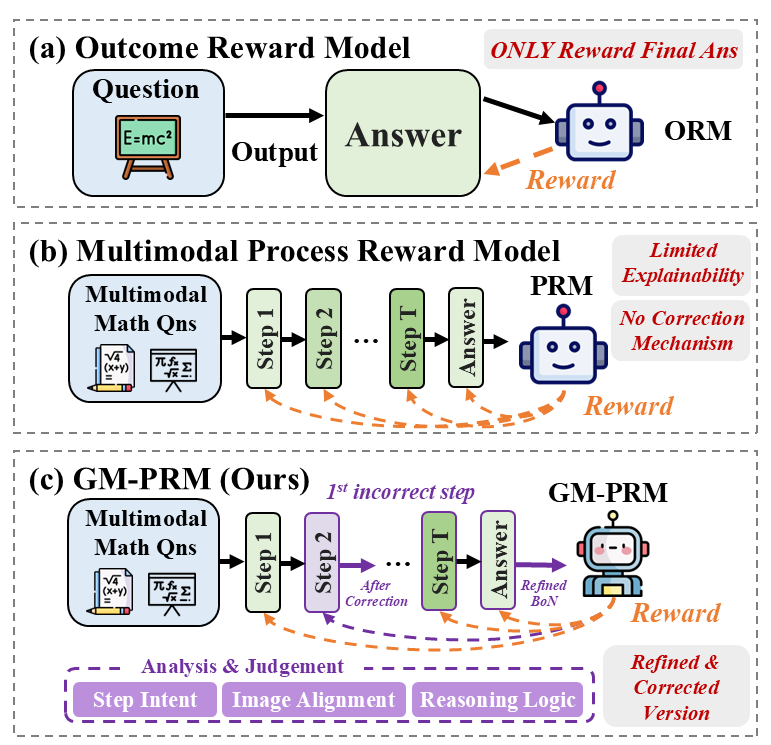}
  \caption{Comparison among ORM (a), PRM (b), and our proposed \model (c) for multimodal math reasoning.}
  \label{fig:comparison}
\end{figure}

To mitigate these reasoning deficiencies, Process Reward Models (PRMs) have emerged as a promising paradigm \cite{gao2024designing,zhong2025comprehensive}. Unlike outcome-based models that only reward correct final answers (shown in Figure \ref{fig:comparison} (a)), PRMs provide fine-grained supervision by evaluating the correctness of each intermediate step in a reasoning chain \cite{zheng2024processbench,lambert2024rewardbench,yan2024errorradar}, as shown in Figure \ref{fig:comparison} (b). This approach has proven effective in the language domain \cite{zeng2025versaprm,yuan2024free,zhang2025openprm}. However, extending PRMs to the multimodal context presents unique challenges \cite{miao2025boosting,du2025mm,li2025devil,cao2025dreamprm}. Existing multimodal PRMs often function as binary classifiers, assigning a simple correct/incorrect label to each step, which offers \textit{limited explanatory power}. Furthermore, they typically possess the ability to identify errors but \textit{lack the mechanism to correct them}, leaving the reasoning process fundamentally broken. This limitation constrains their utility, especially within mechanisms like Best-of-N (BoN) sampling, which remain passive selection processes over a static set of potentially flawed solutions.

In this work, we introduce a novel \textbf{\underline{G}enerative \underline{M}ultimodal \underline{P}rocess \underline{R}eward \underline{M}odel (\model)} to address these limitations, as illustrated in Figure \ref{fig:comparison} (c). Our model transcends the role of a simple verifier and acts as an active reasoning collaborator. Instead of merely outputting a scalar score, our \model leverages its generative capabilities to produce a detailed, interpretable analysis for each reasoning step. This analysis deconstructs the step into three critical aspects: its fundamental \textbf{step intent}, the correctness of its \textbf{image alignment}, and the soundness of its \textbf{reasoning logic}. More importantly, our model is trained not only to identify errors but also to \textbf{generate a refined, corrected version} of the first incorrect step it encounters.

This unique corrective capability enables us to propose a new test-time inference strategy: the \textbf{Refined Best-of-N (Refined-BoN)} process. This dynamic framework moves beyond passive selection by actively enhancing the quality of candidate solutions. When our \model identifies a flawed step within a generated solution, it intervenes by providing a corrected step, which is then used to guide the policy model in generating a new, more promising reasoning trajectory. This iterative refinement process significantly improves the diversity and correctness of the solution pool, leading to a substantial boost in the policy model’s problem-solving performance. Furthermore, we demonstrate that this powerful capability can be achieved with remarkable data efficiency, requiring a significantly smaller training dataset than previous approaches. Our primary contributions are as follows:
\begin{itemize}
% \vspace{-0.2em}
    \item We develop a generative multimodal PRM called \model that \textbf{provides fine-grained, interpretable feedback for mathematical reasoning}. It analyzes each step’s purpose, image alignment, and logical validity, moving beyond simple binary classification to offer deeper insight into the model’s thought process.
    % \vspace{-0.1em}
    \item We introduce a novel Refined-BoN framework that \textbf{leverages the PRM’s generative power to actively correct errors at test time}. It enhances the policy model’s ability to find correct solutions by iteratively improving flawed reasoning paths.
    % \vspace{-0.1em}
    \item We demonstrate the \textbf{effectiveness and data efficiency} of \model, achieving state-of-the-art results on \textbf{multiple multimodal math benchmarks}. Our approach requires only a 20K sample dataset, highlighting the quality of data curation and the power of generative supervision strategy.
    % \vspace{-0.2em}
\end{itemize}

\section{Related Work}
\label{sec:related_work}

\subsubsection{Process Reward Models (PRMs)}
PRMs have been proposed to evaluate the fine-grained step level for model reasoning. During the implementation process, annotating and obtaining a high-quality training dataset incurs a high cost. PRM800K~\cite{lightman2023let} is the first process supervision dataset completely annotated by humans.
To mitigate annotation costs, Math-Shepherd~\cite{wang2023math} proposes Monte Carlo (MC) estimation, while OmegaPRM~\cite{luo2024improve} leverages Monte Carlo Tree Search (MCTS) to automatically evaluate each reasoning step, both utilizing the generation capabilities of Large Language Models (LLMs). Subsequent research has enhanced the effectiveness of PRMs through various methods, including VersaPRM~\cite{zeng2025versaprm}, Implicit PRM~\cite{yuan2024free}, OpenPRM~\cite{zhang2025openprm}, PQM~\cite{li2024process}, PAV~\cite{setlur2024rewarding}, and others. In addition, GenRM~\cite{GenPRM} utilizes the generation ability of reward models to analyze each reasoning step and obtain the score of each step by taking the probability of the special evaluation token. Furthermore, GenPRM~\cite{GenPRM}, ThinkPRM~\cite{khalifa2025process}, R-PRM~\cite{she2025rprm} extend the method of using model generation analysis to evaluate steps to PRMs. 
There are also many studies on benchmarks of PRMs such as ProcessBench~\cite{zheng2024processbench}, PRMBench~\cite{song2025prmbench}, and Socratic-PRMBench \cite{li2025socratic}.

\subsubsection{Multimodal PRMs}
After achieving certain results and progress in the research of language modality in PRMs, research on PRMs has also begun to shift towards multimodal tasks. M-STAR~\cite{liu2024diving} proposes and implements multimodal PRM on multimodal problems. URSA~\cite{luo2025ursa} constructs a dataset by inserting errors and utilizes it to train a multimodal PRM. VisualPRM~\cite{wang2025visualprm} not only uses MC estimation to construct a multimodal VisualPRM400K training dataset, but also proposes a benchmark for multimodal PRMs called VisualProcessBench, which is entirely annotated by humans. Athena-PRM~\cite{wang2025athena} proposes using prediction consistency between strong and weak completers to enhance the quality of automatically annotated data based on MC estimation and improving multimodal PRM by ORM initialization and negative data up-sampling. Moreover, PRM-BAS~\cite{hu2025prm}, MM-PRM~\cite{du2025mm} and DreamPRM~\cite{cao2025dreamprm} also improve the capability of multimodal PRMs. Although several studies have explored multimodal PRMs, applying them to multimodal tasks effectively remains certain challenges, such as insufficient interpretability of the labels assigned to each reasoning step and the inability to correct identified erroneous steps. In our work, we introduce a generative multimodal PRM, \model to solve the above problems.

\begin{figure*}[htb!]
  \centering
  \includegraphics[width=0.7\textwidth]{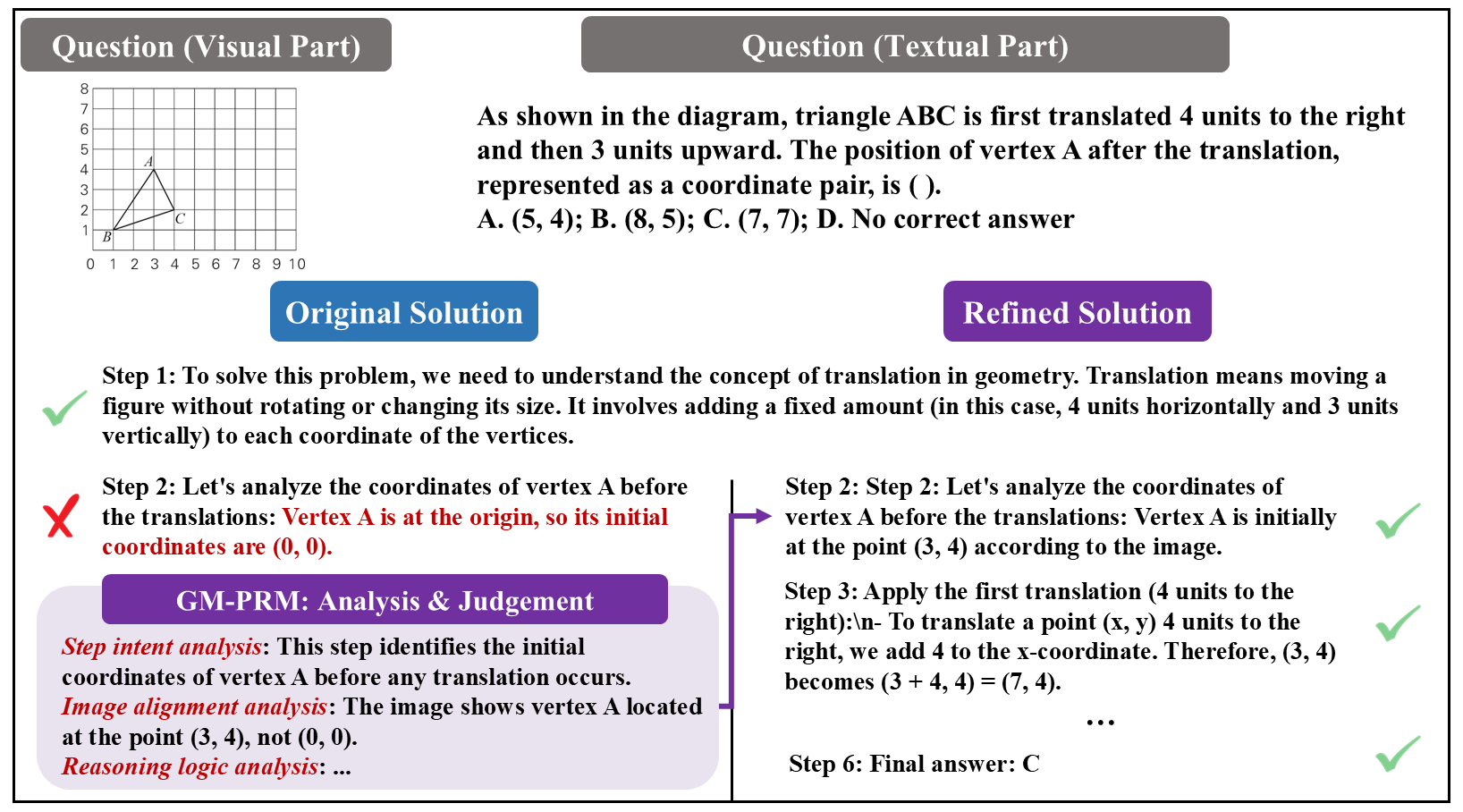}
  \caption{The illustration of a representative example \textit{before} and \textit{after} applying \model. In particular, \model first judges the steps of the original solution generated by the policy model. Subsequently, \model finds that the second step is incorrect and refines the second step to generate the correct version. The correct steps are input to the policy model to generate the refined solution, and finally the correct answer is obtained.}
  \label{fig:case}
\end{figure*}
% \vspace{-2mm}

\section{Methodology}
\label{sec:methodology}
In this section, we first describe how to utilize PRMs and generative PRMs combined with the BoN method to improve the performance of policy models for mathematical problems in Section \ref{sec:prm for math}. Then, we introduce our process to implement multimodal generative PRM, including data construction and model training in Section \ref{data construction}. Finally, we propose a novel Refined-BoN framework for PRMs to enhance its performance beyond traditional BoN method in Section \ref{Refined-BoN}.
\subsection{PRMs for Mathematical Problem}
\label{sec:prm for math}
In this section, we present the implementation methods of PRM and GM-PRM, and provide formal explanations of their usage through mathematical notation.
\subsubsection{Problem and Reasoning Steps Generation}
Let $Q$ denote a mathematical problem. Firstly, an LLM $\pi$ is involved in solving the mathematical problem $Q$. To facilitate reasoning, the problem is combined with a prompt $P$, which includes specific instructions guiding the generation of a step-by-step reasoning process and a final answer. This composite input is then fed into the LLM. When generating a response, $\pi$ generates a sequence of reasoning steps, denoted as $R = \{ r_1, r_2, \ldots, r_T \}$, where $T$ represents the total number of reasoning steps to the given mathematical problem. The above process can be explained as follows:
\begin{equation}
    R = \pi(Q \parallel P),
\end{equation}
where $ \parallel $ denotes the concatenation of the problem $ Q $ and the prompt $ P $, and $ \pi(\cdot) $ represents the  inference of LLM.

\subsubsection{PRM}
A single instance in a training dataset $\mathcal{D}$ to train a PRM comprises three components: (1) a problem statement, (2) a generated response consisting of multiple inference steps, and (3) a corresponding set of binary labels, each taking a value of either 0 or 1, indicating whether the associated reasoning step is incorrect or correct, respectively.

During training, the PRM is optimized using cross-entropy loss and supervised to align its predictions with the ground-truth labels. After being trained, the PRM model is capable of processing new reasoning steps generated by the LLM in response to a given mathematical problem, which means that the PRM is able to assign a scalar score to each individual reasoning step, reflecting the model's confidence in the correctness of each step:
\begin{equation}
f_{\text{PRM}} : (Q, R) \mapsto (s_1, s_2, \dots, s_T),
\end{equation}
where $f_{\text{PRM}} : (\cdot)$ represents the inference of PRM, $ s_i \in [0, 1] $ denotes the confidence score assigned to the $ i $-th reasoning step $ r_i $, and $T$ denotes the number of reasoning steps.

For generative PRM, the binary labels in the training dataset are replaced with textual analyses and judgments, each formulated as a textual choice such as ``incorrect'' or ``correct''. During inference, generative PRM also generates textual critiques and judgments for each step.

\subsubsection{\model}
By extending generative PRMs from the textual modality to a multimodal setting, we introduce \model. In this setting, mathematical problems are represented using both textual and visual information. The input to the policy model comprises the image of the problem, its textual description and task-specific instructions, which are processed jointly to generate reasoning steps. Similarly, during both training and inference, it is essential to provide \model with inputs from both visual and textual modalities, enabling it to perform cross-modal analysis when assigning correctness labels to each reasoning step:
\begin{equation}
f_{\text{\model}} : (Q, I, R) \mapsto (c_1, j_1, \dots, c_T, j_T),
\end{equation}
where $f_{\text{\model}} : (\cdot)$ represents the inference of \model, $ I $ denotes the image of the mathematical problem, $ c_i $ denotes the critique of the $ i $-th reasoning step $ r_i $, and $ j_i $ denotes the textual judgment assigned to the $ i $-th reasoning step $ r_i $.

\subsection{Data Construction}
\label{data construction}
In this section, we present our methodology employed to construct the training data for \model. The process consists of three key stages: (1) the selection of appropriate types and quantities of question data from the VisualPRM400K dataset~\cite{wang2025visualprm}; (2) the generation of textual analysis and judgment data using GPT-4o; and (3) the filtering of the generated data through MC estimation and LLM-as-a-judge techniques to ensure quality and reliability.

\subsubsection{Data Selection}
VisualPRM400K is a large-scale dataset containing approximately 400,000 multimodal process supervision samples. 
In our work, we select plane geometry- and function-related problems from VisualPRM400K to construct a specialized subset and supplement it with corresponding textual analysis for training GM-PRM. This targeted subset with textual critiques supports the effective training of \model, yielding strong performance on geometric and function-based mathematical reasoning tasks. 

\subsubsection{Generation of Analysis and Judgment}
To obtain textual analyses and judgments, we employ GPT-4o to critique each reasoning step from 4 key aspects: \textbf{step intent}, \textbf{image alignment}, \textbf{reasoning logic}, and \textbf{step refinement}.

The aspect of \textbf{step intent} indicates identifying the purpose of each reasoning step. This initial analysis establishes a foundation that allows \model to interpret and evaluate each reasoning step in context more effectively. %By first analyzing the intent behind each step, \model acquires a precise understanding of how each step contributes to the overall reasoning process.
Furthermore, this level of understanding facilitates subsequent error detection and correction tasks, thereby enhancing the overall effectiveness of \model.

The second aspect is \textbf{image alignment}. 
When MLLMs are used for inference in solving multimodal problems, MLLMs often make errors in image alignment, such as misidentifying parallel relationships or incorrectly annotating angles, which leads to flawed solutions. To address this, we employ GPT-4o to produce textual analysis and judgments in image alignment for inference steps, to form the dataset for training \model.

\textbf{Reasoning logic} is an indispensable presence in the step-by-step problem-solving process of MLLMs. However, the occurrences of logical inconsistencies and errors, such as miscalculations and incorrect inferences significantly impact the correctness of the reasoning steps and the final answers. Therefore, it is crucial for \model to be capable of identifying such logical flaws and making accurate judgments regarding the validity of the reasoning logic for each step. In our work, we employ GPT-4o to generate textual analysis and judgments of each step in reasoning logic to form the training dataset. The above process can be formulated as follows:
\begin{equation}
\mathcal{F}: (Q, I, R \parallel P) \mapsto \{SI_i, IA_i, RL_i, FJ_i\}_{i=1}^t,
\end{equation}
where $ \mathcal{F}: (\cdot) $ represents the inference of GPT-4o, $SI_i$ denotes the textual analysis of step intent for the $ i $-th reasoning step, $t$ denotes the number of the first incorrect step or the last step, $1 \leqslant t \leqslant T$, $IA_i = \{IAC_i, IAJ_i\}$ denotes the analysis which contains critique $IAC_i$ and judgment $IAJ_i$ in image alignment of the $ i $-th reasoning step, $RL_i = \{RLC_i, RLJ_i\}$ denotes the analysis which contains critique $RLC_i$ and judgment $RLJ_i$ in image alignment of the $ i $-th reasoning step, $FJ_i$ denotes the final judgment of the $ i $-th reasoning step.

Building on aforementioned three aspects, we further aim for \model to \textbf{correct the first identified erroneous step}. %by leveraging its analysis of step correctness, along with the step intent and the identified causes of error. 
The above information enables \model to generate corrected reasoning steps that are logically coherent, visually accurate, and semantically aligned with the original step intent. The resulting corrected steps can then be used to construct more diverse and accurate inference solutions and ultimately produce more reliable final answers. In our work, we employ GPT-4o to generate a corrected version of the first identified error step in a reasoning process if the first error step is detected to exist by GPT-4o:
\begin{equation}
\mathcal{F}: (Q, I, R \parallel P) \mapsto 
\begin{cases} 
RS, & \text{if incorrect step exists}, \\
\emptyset, & \text{otherwise}.
\end{cases}
\end{equation}
where $RS$ denotes refined step of the first error step in a reasoning process.

In summary, we design a structured prompt for GPT-4o to generate comprehensive analysis data across four dimensions based on the provided problems, associated images, and step-by-step solutions:
\begin{equation}
\mathcal{F}: (Q, I, R \parallel P) \mapsto \mathcal{D},
\end{equation}
where $\mathcal{D}$ denotes the generated training dataset:
\begin{equation}
\mathcal{D} = \{(\{SI_i^k, IA_i^k, RL_i^k, FJ_i^k\}_{i=1}^t, RS^k) \}_{k=1}^K,
\end{equation}
where $k \in \{1,2,\dots,K\}$ represents the $k$-th sample in the dataset, and $K$ denotes the number of the training instances.

\subsubsection{Data Filtering}
The process of constructing training data using GPT-4o can be regarded as an implementation of LLM-as-a-judge methodology. Inspired by the combination of LLM-as-a-judge and MC estimation techniques \cite{zhang2025lessons}, we employ the MC estimation technique proposed by Math-Shepherd \cite{wang2023math} to effectively filter and curate the generated data.

Monte Carlo estimation is a strategy for automated annotation that leverages LLMs or MLLMs to generate multiple subsequent solutions for each step. When applying MC estimation to evaluate a step $r_i$, we use an LLM or an MLLM as a `completer' to finalize multiple subsequent reasoning processes from this step: 
\begin{equation}
f_{comleter} \mapsto \{(r_{i+1}^j, \dots, r_{L_j}^j, a^j)\}_{j=1}^m,
\end{equation}
where $a^j$ is the final answer of the $j$-th finalized solution and $L_j$ is the total number of steps.

Within MC estimation, one type of evaluation method is commonly applied: hard estimation. In hard estimation, a step $r_i$ is deemed correct if at least one subsequent solution reaches the correct final answer $a^*$; otherwise, it is considered incorrect:
\begin{equation}
l_i^{HE} = 
\begin{cases} 
1, & \exists a_j, a_j=a^*, \\
0, & \text{otherwise}.
\end{cases}
\end{equation}

In our data construction process, we employ hard estimation to label the correctness of individual reasoning steps. By integrating LLM-as-a-judge technique and MC estimation, we compare the labels acquired by MC estimation and judgments generated by GPT-4o. Data samples that receive consistent evaluations from both methods are selected as our final training dataset. By integrating these two methods, we aim to further enhance the reliability and quality of the training data, ensuring better performance of \model.

\subsection{Refined-BoN Process}
\label{Refined-BoN}
When applying Test-time Scaling (TTS) for LLMs and MLLMs, a widely adopted method is Best-of-N (BoN) approach. In the BoN process, a policy model is employed to generate N candidate solutions, which are then evaluated by reward models or self-consistency to select the optimal solution. However, during the BoN process, policy models are under identical prompting conditions when generating multiple solutions, which leads to the problem that the solutions often lack diversity and may exhibit limited correctness. In our work, we propose a novel Refined-BoN framework utilizing TTS techniques to enhance the diversity and accuracy of generated solutions, thereby improving the reasoning capabilities of policy models.

\subsubsection{Refined-BoN Method}
As shown in Figure \ref{fig:case}, in Refined-BoN process, we first employ an MLLM as the policy model to generate $N/2$ initial solutions to a multimodal problem, and then these solutions are evaluated step-by-step by \model. For the subsequent $N/2$ solutions, the policy model generates them under varying conditions, informed by the evaluation of the preceding $N/2$ solutions: If \model identifies an incorrect reasoning step within a solution, it stops evaluating and refines the first erroneous step by generating a corrected version. This corrected step, along with all previously validated correct steps, is then input back into the policy model to continue the solution generation process. Conversely, if \model determines that all steps in a particular solution are correct, we employ the policy model to generate a new solution using the same prompt. Through this regeneration mechanism, we obtain the additional $N/2$ solutions. Subsequently, we employ \model to evaluate the subsequent $N/2$ solutions.

\subsubsection{Solution Selection}
After applying the Refined-BoN process, we obtain $N$ solutions for each problem, each accompanied by step-level correctness judgments. Moreover, we divide all the solutions into two categories: one where \model judges that it contains incorrect steps, and the other where \model judges that all its steps are correct. Furthermore, we take the probability of \model generating the associated ``Correct'' and ``Incorrect'' tokens as the score of each step.
%for each reasoning step, and these special tokens serve as indicators of the validity of individual steps within the generated solutions.

Among the $N$ generated solutions, if there exist solutions in which all reasoning steps are judged correct, we calculate the average of the scores of all steps in these solutions as the overall score of the solution, and select the solution with the highest average score as the optimal solution. 

For N solutions to a problem, if \model determines that all $N$ solutions contain incorrect steps, we calculate the average score of all steps in a solution as the overall score of the solution, and select the solution with the highest overall score as the final answer.

\section{Experiments}
\begin{table*}[htb!]
\centering
\tiny
\renewcommand\tabcolsep{3pt} % 减小列间距
\renewcommand\arraystretch{1.0} % 微调行高
\resizebox{0.7\linewidth}{!}{
    \begin{tabular}{
        @{}
        c 
        | >{\centering\arraybackslash}m{0.8cm} 
        | >{\centering\arraybackslash}m{0.8cm} 
        | >{\centering\arraybackslash}m{0.8cm} 
        | >{\centering\arraybackslash}m{0.8cm} 
        | >{\centering\arraybackslash}m{0.8cm} 
        | >{\centering\arraybackslash}m{0.8cm}
        @{}
    }
    \shline
    \multirow{1}{*}{\textbf{MLLMs}} &
    \multicolumn{1}{c|}{\textbf{MathVista}} &
    \multicolumn{1}{c|}{\textbf{MathVision}} &
    \multicolumn{1}{c|}{\textbf{MathVerse}} &
    \multicolumn{1}{c|}{\textbf{DynaMath}} &
    \multicolumn{1}{c|}{\textbf{WeMath}} &
    \multicolumn{1}{c}{\textbf{Average}}
    \\ 
    \midrule
  
    {MiniCPM-V2.6-8B}  & 44.3 & 16.0 & 18.9 & 22.6 & 38.6 & 28.1\\
    \hfill~\textbf{+ \model (Ours)} & \textbf{51.0} &\textbf{18.1} & \textbf{24.4} &\textbf{25.7} & \textbf{51.0} & \textbf{34.0} \\
    \hfill~\textbf{Improvements}& \underline{+6.7} & \underline{+2.1} & \underline{+5.5} & \underline{+3.1} & \underline{+12.4} & \underline{+5.9} \\
    \midrule

    {Llama-3.2-11B-Vision}  & 44.5 & 14.3 & 16.5 & 28.4 & 46.1 & 30.0\\
    \hfill~\textbf{+ \model (Ours)} & \textbf{49.5} &\textbf{18.2} & \textbf{18.8} &\textbf{32.7} & \textbf{53.4} & \textbf{34.5} \\
    \hfill~\textbf{Improvements}& \underline{+5.0} & \underline{+3.9} & \underline{+2.3} & \underline{+4.3} & \underline{+7.3} & \underline{+4.5} \\
    \midrule

    {Qwen2.5-VL-7B}  & 63.2 & 25.1 & 32.8 & 35.0 & 60.6 & 43.3\\
    \hfill~\textbf{+ \model (Ours)} & \textbf{65.0} &\textbf{28.2} & \textbf{37.4} &\textbf{39.2} & \textbf{69.0} & \textbf{47.8} \\
    \hfill~\textbf{Improvements}& \underline{+1.8} & \underline{+3.1} & \underline{+4.6} & \underline{+4.2} & \underline{+8.4} & \underline{+4.5} \\
    \midrule

    {InternVL3-8B}  & 50.6 & 20.3 & 25.0 & 27.0 & 50.9 & 34.8\\
    \hfill~\textbf{+ \model (Ours)} & \textbf{55.7} &\textbf{22.2} & \textbf{31.7} &\textbf{33.4} & \textbf{59.2} & \textbf{40.4} \\
    \hfill~\textbf{Improvements}& \underline{+5.1} & \underline{+1.9} & \underline{+6.7} & \underline{+6.4} & \underline{+8.3} & \underline{+5.6} \\
    \midrule

    {InternVL3-38B}  & 68.3 & 34.9 & 37.8 & 40.1 & 66.4 & 49.5\\
    \hfill~\textbf{+ \model (Ours)} & \textbf{69.9} &\textbf{37.0} & \textbf{39.1} &\textbf{43.1} & \textbf{72.9} & \textbf{52.4} \\
    \hfill~\textbf{Improvements}& \underline{+1.6} & \underline{+2.1} & \underline{+1.3} & \underline{+3.0} & \underline{+6.5} & \underline{+2.9} \\
    \midrule

    {InternVL3-78B}  & 68.0 & 34.6 & 36.0 & 38.1 & 65.7 & 48.5\\
    \hfill~\textbf{+ \model (Ours)} & \textbf{70.7} &\textbf{37.1} & \textbf{40.6} &\textbf{39.9} & \textbf{72.2} & \textbf{52.1} \\
    \hfill~\textbf{Improvements}& \underline{+2.7} & \underline{+2.5} & \underline{+4.6} & \underline{+1.8} & \underline{+6.5} & \underline{+3.6} \\

    \shline
    \end{tabular}
}
% \vspace{-2mm}
\caption{Percentage accuracy scores (\%) of multiple MLLMs across five datasets. For each MLLM, the first row shows the baseline, the second shows the final result with \model, and the third shows the improvement. Only positive improvements are \underline{underlined}. The best results are highlighted in \textbf{bold}. All values are reported after rounding to three decimal places.}
\label{tab:main result}
\end{table*}

\label{sec:experiment}
In this section, we introduce our experimental setup to assess \model under the Refined-BoN process on five multimodal mathematical benchmarks in Section \ref{sec:experimental setup}. In addition, we present the results of our experiments and three conclusions analyzed from the results in Section \ref{sec:main results}. Finally, we show the ablation studies in Section \ref{sec:ablation study}.

\subsection{Experimental Setup}
\label{sec:experimental setup}
\subsubsection{Benchmarks.}
We evaluate \model across five datasets, including MathVista \cite{lu2023mathvista}, MathVision \cite{wang2024measuring}, MathVerse \cite{zhang2024mathverse}, DynaMath \cite{zou2024dynamath} and WeMath \cite{qiao2024we}. The datasets contain diverse problem types, such as plane geometry, functions, puzzle tests, \textit{etc}. We use Vision-Only subset of MathVerse dataset and Plane-Geometry subset of DynaMath.
\subsubsection{Settings.}
We employ \model as the critic model for Refined-BoN evaluation and set N to 8 by default. For MLLMs, we select six models as the policy models to generate step-by-step reasoning processes. When reasoning, we set the temperature of the policy models to 0.7 and top-p to 0.9. For comparison, we use the average accuracy of N sets of answers generated by policy models as baselines.
\subsubsection{Training Details.}
To train \model, we use Qwen2.5-VL-7B-Instruct as our base model and perform supervised fine-tuning (SFT) with all parameters trainable except for the frozen Vision Transformer (ViT) encoder. During the training process, we utilize bfloat16 mixed-precision and DeepSpeed with zero3 technology and set the training consists of 2 epochs. For batch size, the batch size on each training device is set to 2, and through gradient accumulation, the effective batch size is extended to 16. Moreover, we use two A800 GPUs to train \model, and the AdamW optimizer is used with an initial learning rate of $1\times10^{-5}$. The learning rate schedule involves a linear warm-up with the warm-up ratio equal to 0.05 followed by linear decay.

\subsection{Main Results}
\label{sec:main results}
As shown in Table \ref{tab:main result}, integrating \model with the Refined-BoN process consistently improves performance across five benchmark datasets for six different MLLMs. On average, our method yields notable accuracy gains, with improvements of +5.9 for MiniCPM-V2.6-8B, +4.5 for Llama-3.2-11B-Vision, +4.5 for Qwen2.5-VL-7B, and +5.6 for InternVL3-8B. 

A closer look at dataset-level results reveals that the improvements are not uniform. The WeMath benchmark shows the most significant enhancement, with MiniCPM-V2.6-8B improving by +12.4 points, highlighting the ability of our method to strengthen mathematical reasoning on challenging problems. Similarly, MathVerse and DynaMath exhibit consistent gains of +4.5–6.7 points across multiple models, suggesting that our approach particularly benefits datasets requiring complex symbolic manipulation and multi-step reasoning. In contrast, MathVision improvements are more modest (+1.9–3.9), indicating that the visual reasoning component may already be relatively strong in baseline models.

\begin{figure}[h]
  \centering
  \includegraphics[width=0.43\textwidth]{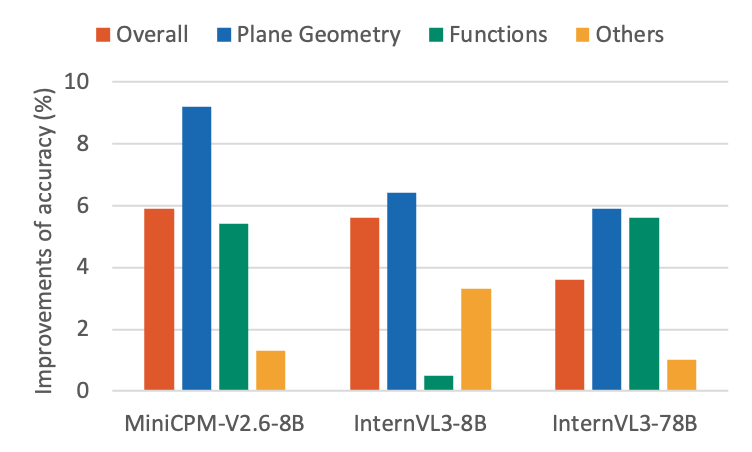}
  \caption{Improvements of the average percentage accuracy (\%) of multiple MLLMs across different question types in MathVista, MathVision and MathVerse datasets.}
  \label{fig:conclusion2}
\end{figure}

\textbf{\model combined with the Refined-BoN process demonstrates strong generalization across diverse multimodal mathematical problems, with particularly remarkable gains in plane geometry tasks.} As illustrated in Figure \ref{fig:conclusion2}, even after excluding plane geometry and function problems, policy models still achieve notable improvements across the datasets. This indicates that although \model is primarily trained on a dataset composed of plane geometry and function problems, it generalizes effectively to other types of multimodal mathematical problems. Moreover, as shown by the averaged results in Figure \ref{fig:conclusion2}, the improvements achieved by \model with Refined-BoN on plane geometry problems consistently exceed those on the overall dataset, function problems, and other categories, underscoring the exceptional effectiveness of our method in tackling plane geometry tasks.

% \vspace{-2mm}

\textbf{The Refined-BoN process yields disproportionately larger gains for models with lower baseline performance.} As shown in Table \ref{tab:main result}, InternVL3-38B starts with the highest initial average accuracy among all policy models (49.5\%) and achieves a modest improvement of +2.9 points (+5.9\%). In contrast, Qwen2.5-VL-7B, which has the highest baseline accuracy (43.3\%) among models with fewer than 12 billion parameters, improves by +4.5 points (+10.4\%), surpassing the relative gains of InternVL3-38B. Notably, MiniCPM-V-2.6-8B demonstrates the most significant relative improvement, achieving +5.9 points (+21.0\%), despite its lower initial score. These results suggest that models with weaker baseline performance benefit more from the refinement mechanism of \model with Refined-BoN, likely because the process effectively corrects errors in reasoning steps, leaving greater room for improvement.

\begin{figure*}[t!]
    \centering
    \begin{minipage}{0.32\linewidth}
        \centering
        \includegraphics[width=\linewidth]{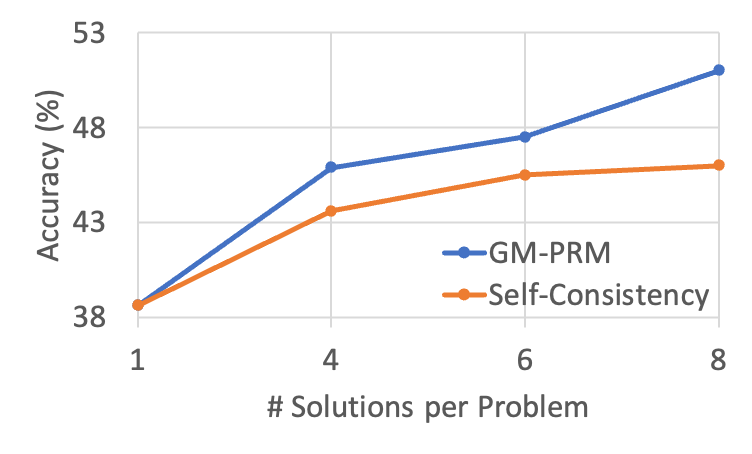}
        \caption*{(a) MiniCPM-V-2.6-8B }
    \end{minipage}
    \hfill
    \begin{minipage}{0.32\linewidth}
        \centering
        \includegraphics[width=\linewidth]{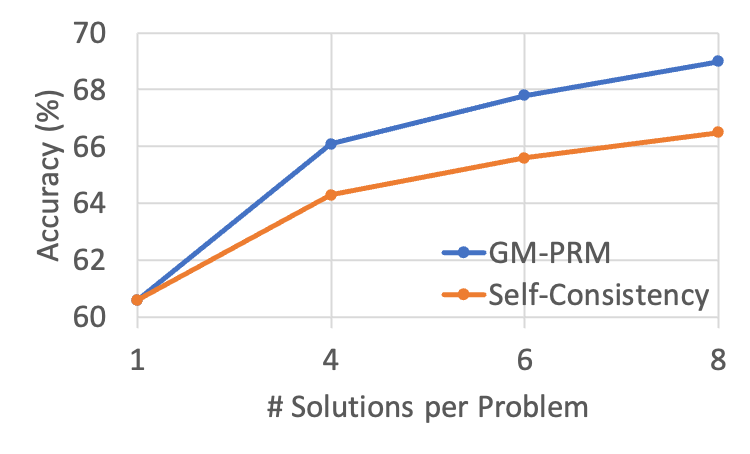}
        \caption*{(b) Qwen2.5-VL-7B }
    \end{minipage}
    \hfill
    \begin{minipage}{0.32\linewidth}
        \centering
        \includegraphics[width=\linewidth]{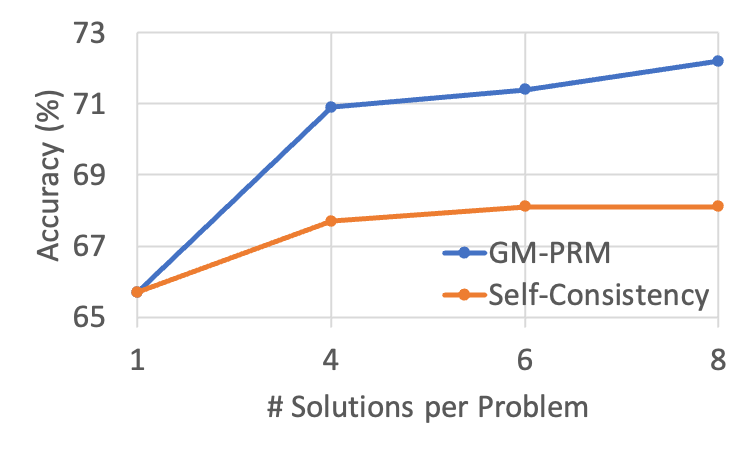}
        \caption*{(c) InternVL3-78B }
    \end{minipage}
    \caption{The results of changing the value of N in the Refined-BoN process on the WeMath across different policies. As N increases, the effectiveness of \model in enhancing accuracy improves and surpasses that of Self-Consistency.}
    \label{fig:BoN}
\end{figure*}

\subsection{Hyperparameter \& Ablation Study}
\label{sec:ablation study}
\subsubsection{Number of solution samples N in Refined BoN.} 
Following Test-time Scaling technique, we vary the number of N in the Refined-BoN process to evaluate the performance of \model in comparison to the Self-Consistency baseline.

Figure \ref{fig:BoN} depicts WeMath accuracy as the number of sampled solutions per problem (\(N\)) increases from 1 to 8. Across all three backbones—MiniCPM‑V‑2.6‑8B, Qwen2.5‑VL‑7B, and InternVL3‑78B—both \model and the Self‑Consistency (SC) baseline benefit from a larger sampling budget, yet \model exhibits a noticeably steeper growth curve.  

Under the widely adopted Best‑of‑8 setting, \model delivers gains of 4.9 and 3.5 over SC on MiniCPM‑V‑2.6‑8B and Qwen2.5‑VL‑7B, respectively. Even for the 78B‑parameter InternVL3, \model maintains a substantial 4.1 margin. These results indicate that the proposed refinement strategy not only scales to larger models but also converts additional candidate solutions into accuracy more effectively than Self‑Consistency, thereby underscoring the robustness and versatility of \model.

Furthermore, for MiniCPM-V-2.6-8B, \model surpasses the self-consistency baseline by 2.1, 2.2, and 4.9 points under the Best-of-4, Best-of-6, and Best-of-8 settings, respectively, indicating a steadily increasing performance gap between \model and self-consistency as N increases.

\subsubsection{Methods for aggregating step scores.}
For PRMs, the method used to aggregate step scores into an overall solution score plays a critical role. In this part, we compare several different aggregation strategies, including averaging step scores, selecting the maximum step score, and selecting the minimum step score. Since step-by-step solutions that contain steps judged incorrect are often not evaluated or scored for all steps, this experiment focuses exclusively on solutions where all steps are judged correct.
\begin{figure}[htbp]
  \centering
  \includegraphics[width=0.44\textwidth]{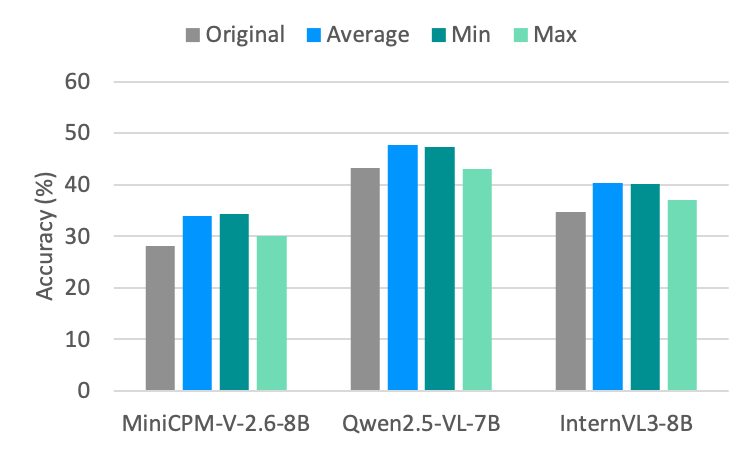}
  \caption{Average percentage accuracy (\%) of MLLMs via different aggregation methods across five datasets.}
  \label{fig:Aggregation}
\end{figure}
% \vspace{-2mm}

The results are illustrated in the Figure \ref{fig:Aggregation}. Across all policy models and datasets, we find that both averaging the step scores and selecting the minimum score significantly outperform the strategy of selecting the maximum score. This suggests that either the average or the minimum score provides a more accurate reflection of the overall quality of a solution than the maximum score. Between the minimum and average aggregation methods, we observe that averaging performs slightly better. This improvement may stem from the fact that the average score takes into account all problem-solving steps, providing a more comprehensive evaluation, whereas the minimum score reflects only the step with the lowest score and thus offers a less holistic assessment.

\subsubsection{Refined-BoN vs. BoN.}
The Refined-BoN process aims to enhance the diversity of N candidate solutions by refining the steps judged incorrect and integrating the refined steps with the steps judged correct into the prompt for the policy models. In this part, we use the Pass@k metric to evaluate the diversity and accuracy of policy models in generating multiple solutions to the given problems.

The results are summarized in the Table \ref{tab:Pass@8}. Overall, the Refined-BoN process improves Pass@8 scores compared to the standard BoN process across multiple policy models and five benchmark datasets. Specifically, it increases the average Pass@8 values of MiniCPM-V-2.6-8B, Llama-3.2-11B-Vision, and InternVL3-8B by 0.9, 1.3, and 0.9 points, respectively, across the five datasets, demonstrating the effectiveness of the Refined-BoN approach in enhancing the diversity and correctness of the generated solutions.

\begin{table}[h!]
  \centering
  \small
  \tabcolsep=3mm
    \begin{tabular}{l|c|c|c}
    \shline
    \textbf{MLLMs} & \textbf{BoN} & \textbf{Refined-BoN} & \textbf{Diff.} \\
    \hline \hline
    MiniCPM-V-2.6-8B & 62.5 & \textbf{63.4} & \underline{+0.9} \\
    Llama-3.2-11B-Vision & 62.7 & \textbf{64.0} & \underline{+1.3} \\
    InternVL3-8B & 65.3 & \textbf{66.2} & \underline{+0.9} \\
    \shline
    \end{tabular}%
    \caption{Average percentage Pass@8 scores of BoN and Refined-BoN across five datasets for different models.}
  \label{tab:Pass@8}%
  % \vspace{-0.5em}
\end{table}%

\section{Conclusion}
\label{sec:conclusion}

In this work, we introduced \model, a novel paradigm that transforms the reward model from a passive judge into an active reasoning collaborator for multimodal mathematics. By providing fine-grained, interpretable analysis and, more critically, generating corrections for erroneous steps, \model moves beyond simple binary verification. This unique corrective capability powers our Refined Best-of-N (Refined-BoN) framework, which actively improves flawed reasoning trajectories at test time. Our experiments demonstrate that this approach achieves state-of-the-art results on multiple benchmarks, significantly boosting policy model performance with remarkable data efficiency. The consistent gains across diverse models and problem types underscore the robustness and generalizability of our method. This shift from passive error detection to generative, collaborative correction represents a crucial advance in multimodal reasoning.

\nobibliography*

% \section{Acknowledgments}
% pending
\bigskip

% \clearpage
\bibliography{aaai2026}

\newpage

\appendix
\section{Appendix}
\subsection{More Related Work} 
\subsubsection{Multimodal Large Language Models (MLLMs)}
The advancement of artificial intelligence has advanced the development of Multimodal Large Language Models (MLLMs). MLLMs extend the capabilities of language-centric models by integrating multiple sensory inputs, primarily visual and auditory, with text. Unlike traditional Large Language Models (LLMs) which process solely textual data, MLLMs are designed to perceive and reason across modalities such as vision and language, thereby achieving the fusion and interaction of multimodal information. 
The development of MLLMs has been driven by extensive efforts, including enhancements across model structure and data curation. %, and training algorithms. 
In terms of model structure, multiple studies \cite{bai2025qwen25vltechnicalreport, liu2023interngptsolvingvisioncentrictasks, yao2024minicpmvgpt4vlevelmllm} achieve notable performance through a method that utilizing connectors to align the embeddings of vision from Vision Foundation Models (VFMs) \cite{chen2024internvlscalingvisionfoundation} with the latent space of LLMs \cite{bai2023qwentechnicalreport, touvron2023llamaopenefficientfoundation, touvron2023llama2openfoundation}. Alternatively, another line of research \cite{grattafiori2024llama3herdmodels, tian2024mminterleavedinterleavedimagetextgenerative} enhances pre-trained LLMs by adding supplementary layers to integrate visual features, which reduces the number of visual tokens but incurs additional training costs. 
Regarding dataset curation, recent research has achieved substantial advancements. Specifically, MultimodalC4 \cite{zhu2023multimodalc4openbillionscale} extends C4 corpus containing only text with images and constructs a corpus that supports pre-training for MLLMs. Furthermore, OmniCorpus \cite{li2024omnicorpusunifiedmultimodalcorpus} delivers a large-scale yet noisy multimodal dataset suitable for pre-training, and MMInstruct \cite{Liu_2024} presents an open-source collection of high-quality data designed for instruction tuning.
The majority of research efforts have been concentrated on the training processes of MLLMs, leaving significant room for exploration in Test-Time Scaling (TTS) technique. In our work, we investigate the potential of enhancing the performance of MLLMs by incorporating Process Reward Model (PRM) into the TTS framework.

\subsection{Benchmark}
We provide more details about the Refined-BoN test benchmarks in Table \ref{tab:benchmark}:

\begin{table}[h!]
  \centering
  \small
  \tabcolsep=3mm
    \begin{tabular}{l|c|c}
    \shline
    \textbf{Benchamrks} & \textbf{Split} & \textbf{\# Sample} \\
    \hline \hline
    DynaMath & Plane Geometry & 770 \\
    MathVerse & Vision-Only & 788 \\
    MathVista & Testmini & 1000 \\
    WeMath & Testmini & 1740 \\
    MathVision & Full & 3040 \\    
    \shline
    \end{tabular}%
    \caption{More details about the Refined-BoN test benchmarks.}
  \label{tab:benchmark}%
  % \vspace{-0.5em}
\end{table}%

\subsection{Dataset}
To ensure a balanced distribution of process labels, we carefully construct the training dataset. The final dataset used to train \model contains 19,614 samples in total, comprising 9,061 solutions that contain incorrect steps—as jointly identified by GPT-4o and Monte Carlo (MC) estimation—and 10,553 solutions in which all steps are judged to be correct.

\subsection{Prompt}
\begin{figure}[htb!]
  \centering
  \includegraphics[width=0.47\textwidth]{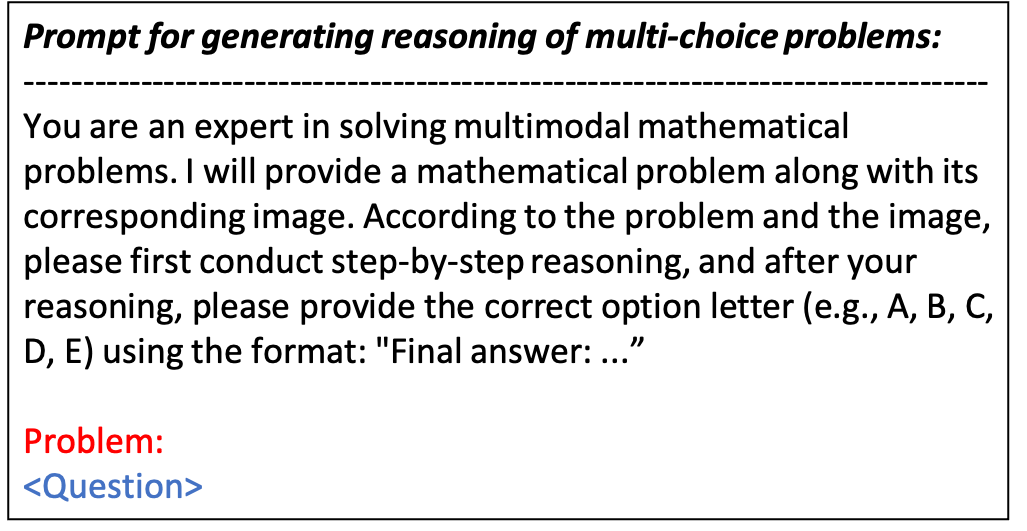}
  \caption{Prompt for policy models to generate reasoning and final answers of multi-choice problem.}
  \label{fig:prompt_1}
\end{figure}

In this section, we introduce the prompts used to construct the training dataset and generate the reasoning processes and final answers. 
The prompt we guide the policy models to generate reasoning processes and final answers of multi-choice problems is represented in Figure \ref{fig:prompt_1}.

\begin{figure}[htb!]
  \centering
  \includegraphics[width=0.47\textwidth]{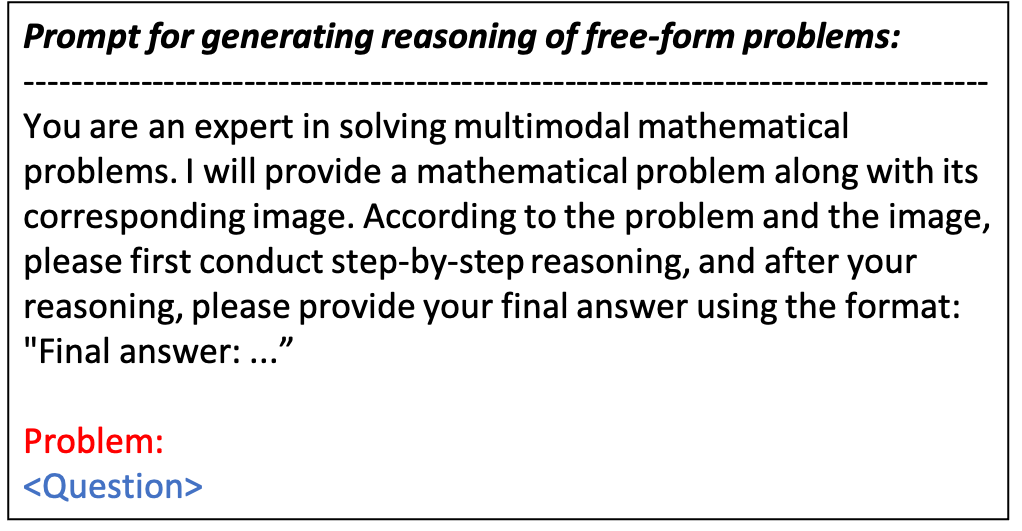}
  \caption{Prompt for policy models to generate reasoning and final answers of free-form problem.}
  \label{fig:prompt_2}
\end{figure}

The prompt we guide the policy models to generate reasoning processes and final answers of free-form problems is represented in Figure \ref{fig:prompt_2}.

\begin{figure*}[htb!]
  \centering
  \includegraphics[width=1.0\textwidth]{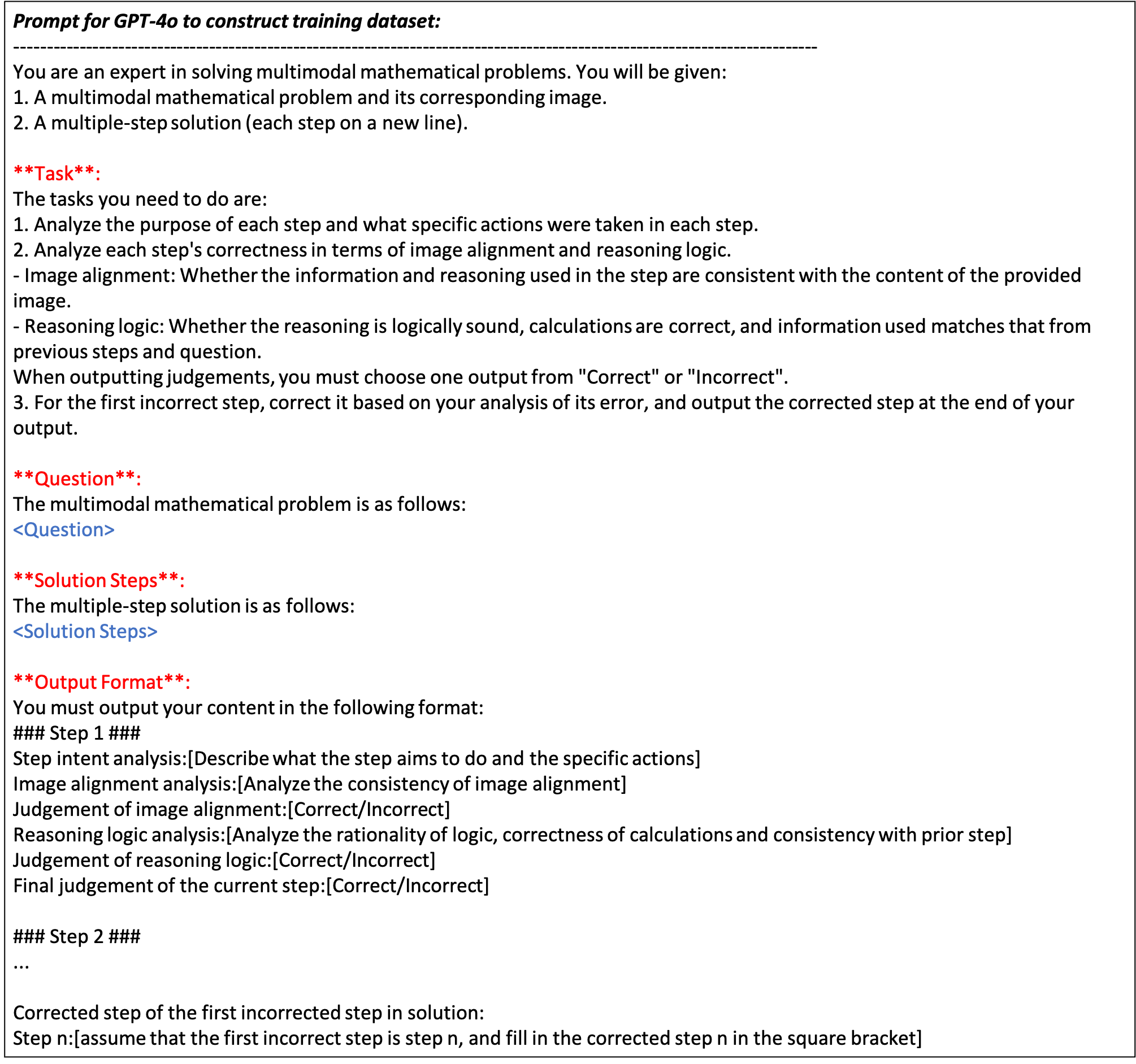}
  \caption{Prompt for GPT-4o to construct training dataset.}
  \label{fig:prompt}
\end{figure*}

The prompt we use to employ GPT-4o to generate the training dataset is shown in Figure \ref{fig:prompt}. 

\subsection{More Cases}
\begin{figure*}[htb!]
  \centering
  \includegraphics[width=1.0\textwidth]{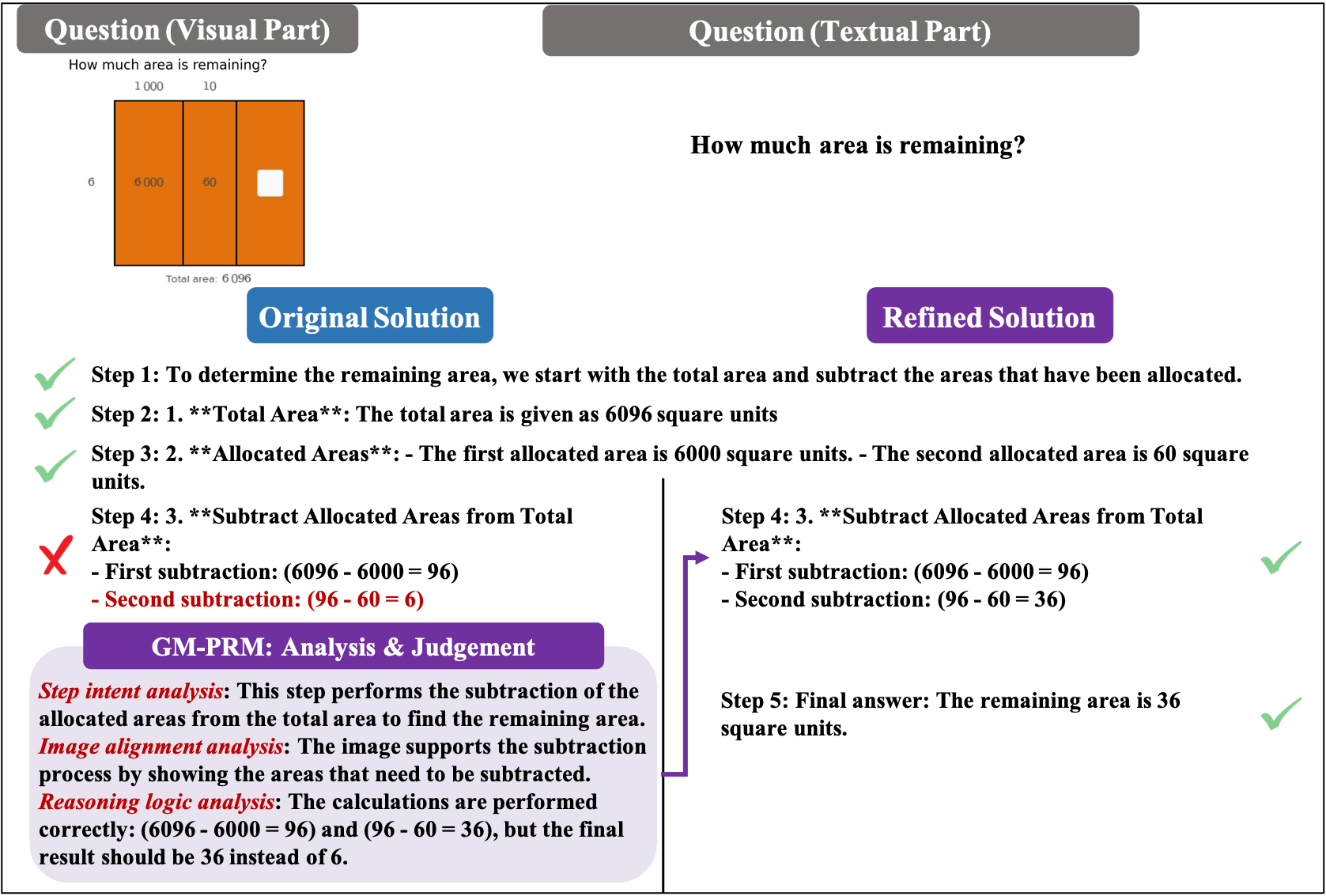}
  \caption{The illustration of a representative example \textit{before} and \textit{after} applying \model. In particular, \model first judges the steps of the original solution generated by the policy model. Subsequently, \model finds that the forth step is incorrect regarding reasoning logic and refines the forth step to generate the correct version. The correct steps are input to the policy model to generate the refined solution, and finally the correct answer is obtained.}
  \label{fig:more_case_1}
\end{figure*}

\begin{figure*}[htb!]
  \centering
  \includegraphics[width=1.0\textwidth]{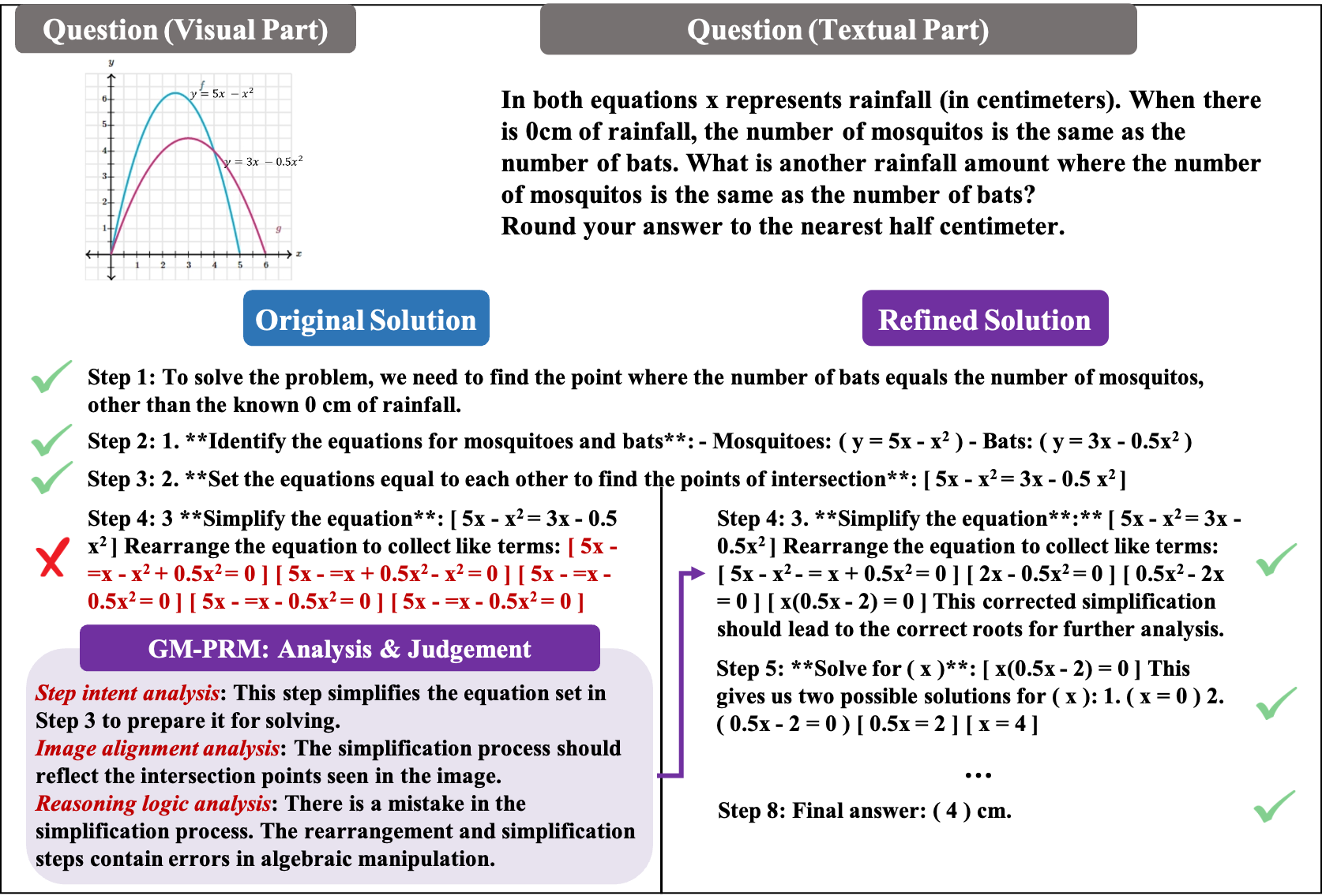}
  \caption{The illustration of a representative example \textit{before} and \textit{after} applying \model. In particular, \model first judges the steps of the original solution generated by the policy model. Subsequently, \model finds that the forth step is incorrect regarding reasoning logic and refines the forth step to generate the correct version. The correct steps are input to the policy model to generate the refined solution, and finally the correct answer is obtained.}
  \label{fig:more_case_2}
\end{figure*}

We collect 2 more cases of \model and the Refined-BoN process illustrated in Figure \ref{fig:more_case_1} and \ref{fig:more_case_2}.

\end{document}